\documentclass[pdflatex, sn-basic, Numbered]{sn-jnl}

\usepackage{graphicx}%
\usepackage{multirow}%
\usepackage{amsmath,amssymb,amsfonts}%
\usepackage{amsthm}%
\usepackage{mathrsfs}%
\usepackage[title]{appendix}%
\usepackage{xcolor}%
\usepackage{textcomp}%
\usepackage{textgreek}
\usepackage{manyfoot}%
\usepackage{array}
\usepackage{booktabs}%
\usepackage{algorithm}%
\usepackage{algorithmicx}%
\usepackage{algpseudocode}%
\usepackage{listings}%
\usepackage[compatibility=false]{caption}
\usepackage{subcaption} %
\usepackage{longtable}%
\usepackage{verbatimbox}%
\usepackage{makecell}%
\usepackage{hyperref}
\usepackage{tabularx}
\usepackage{threeparttable}

\theoremstyle{thmstyleone}%
\theoremstyle{thmstyletwo}%

\theoremstyle{thmstylethree}%

\newcolumntype{L}[1]{>{\raggedright\arraybackslash}p{#1}}
\usepackage{longtable}
\usepackage{booktabs}

\begin{document}

\title[Article Title]{Evaluating Hydro-Science and Engineering Knowledge of 
Large Language Models}


\author[1]{\fnm{Shiruo} \sur{Hu}}
\author[1]{\fnm{Wenbo} \sur{Shan}}
\author[1]{\fnm{Yingjia} \sur{Li}}
\author[1]{\fnm{Zhiqi} \sur{Wan}}
\author[1]{\fnm{Xinpeng} \sur{Yu}}
\author[2]{\fnm{Yunjia} \sur{Qi}}
\author[2]{\fnm{Haotian} \sur{Xia}}
\author[3]{\fnm{Yang} \sur{Xiao}}
\author[3]{\fnm{Dingxiao} \sur{Liu}}
\author[1]{\fnm{Jiaru} \sur{Wang}}
\author[1]{\fnm{Chenxu} \sur{Gong}}
\author[1]{\fnm{Ruixi} \sur{Zhang}}
\author[1]{\fnm{Shuyue} \sur{Wu}}
\author[1]{\fnm{Shibo} \sur{Cui}}
\author[1]{\fnm{Chee Hui} \sur{Lai}}
\author[1,4]{\fnm{Wei} \sur{Luo}}
\author[4]{\fnm{Yubin} \sur{He}}
\author[2]{\fnm{Bin} \sur{Xu}}
\author*[1]{\fnm{Jianshi} \sur{Zhao}}\email{zhaojianshi@tsinghua.edu.cn}

\affil*[1]{\orgdiv{State Key Laboratory of Hydro-Science and Engineering, Department of Hydraulic Engineering}, \orgname{Tsinghua University}, \orgaddress{\city{Beijing}, \postcode{100084}, \country{China}}}

\affil[2]{\orgdiv{Department of Computer Science and Technology}, \orgname{Tsinghua University}, \orgaddress{\city{Beijing}, \postcode{100084}, \country{China}}}

\affil[3]{\orgname{Zhipu AI}, \orgaddress{\city{Beijing}, \postcode{100084}, \country{China}}}

\affil[4]{\orgname{CHN Energy Dadu River Big Data Services Co., Ltd.}, \orgaddress{\city{Chengdu}, \postcode{610041}, \state{Sichuan}, \country{China}}}

\abstract{Hydro-Science and Engineering (Hydro-SE) is a critical and irreplaceable domain that secures human water supply, generates clean hydropower energy, and mitigates flood and drought disasters. Featuring multiple engineering objectives, Hydro-SE is an inherently interdisciplinary domain that integrates scientific knowledge (e.g., hydrology, meteorology) with engineering expertise (e.g., hydraulic structures, geotechnical engineering). This integration necessitates extensive expert collaboration in decision-making, which poses difficulties for intelligence and leads to operational inefficiencies. With the rapid advancement of large language models (LLMs), their potential to enhance the intelligence of Hydro-SE is being increasingly explored. However, the knowledge and application abilities of LLMs in Hydro-SE have not been sufficiently evaluated, which is important to ensure their reliability of practical deployment in real-world scenarios. To address this issue, we propose the Hydro-SE LLM evaluation benchmark (Hydro-SE Bench), which contains 4,000 multiple-choice questions. Hydro-SE Bench covers nine subfields and enables evaluation of LLMs in aspects of basic conceptual knowledge, engineering application ability, and reasoning and calculation ability. Based on Hydro-SE Bench, we evaluated 10 SOTA large-parameter commercial LLMs and 6 small-parameter open-source LLMs. The results show that the accuracy values vary among 0.74–0.80 for commercial LLMs, and among 0.41–0.68 for small-parameter LLMs. While LLMs perform well in subfields closely related to natural and physical sciences, they struggle with domain-specific knowledge such as industry standards and hydraulic structures. Our findings indicate that model scaling mainly improves reasoning and calculation abilities, but there is still great potential for LLMs to better handle problems in practical engineering application. To the best of our knowledge, Hydro-SE Bench is the first benchmark that comprehensively evaluates LLMs' knowledge and capabilities across different subfield dimensions. This study highlights the strengths and weaknesses of LLMs for Hydro-SE tasks, providing model developers with clear training targets and Hydro-SE researchers with practical guidance for applying LLMs.}




\maketitle

\section{Introduction}\label{sec1}

Hydro-Science and Engineering (Hydro-SE) serves as the critical foundation supporting sustainable water resources management, flood and drought mitigation, hydraulic infrastructure construction, and river–ecosystem restoration. As global climate variability intensifies and water-related extremes become more frequent, the demand for accurate hydrology forecast, resilient infrastructure design, and integrated basin-scale management continues to grow. However, the inherently interdisciplinary nature of Hydro-SE requires knowledge spanning hydrology, hydraulics, geotechnical engineering, and engineering safety, placing substantial demands on expert-driven decision-making processes, often resulting in high labor requirements and limited operational efficiency. With increasing challenges in water issues worldwide, improving efficiency, engineering safety, and the level of intelligence in the Hydro-SE domain has become a pressing research priority \citep{RN8}.

The rapid advancement of artificial intelligence, particularly large language models (LLMs), has created great opportunities for promoting efficiency and enhancing intelligence in the Hydro-SE domain. Numerous studies have demonstrated LLMs’ remarkable capabilities in knowledge integration \citep{RN9}, complex reasoning \citep{RN10, RN11, RN14}, and autonomous agentic behavior \citep{RN12, RN13}. Building on these advances, LLMs have been applied across various domains such as finance \citep{RN15, RN24}, law \citep{RN20}, and chemistry \citep{RN22, RN23}, confirming their potential to automate tasks and enhance efficiency. The expanding evidence of LLMs’ effectiveness across multiple domains has inspired growing interest in their application to Hydro-SE among both researchers and industry practitioners \citep{RN25, RN26}. However, excelling in the Hydro-SE domain requires expertise across a wide range of subfields \citep{RN28}, such as hydrology, river dynamics, and hydraulic structure design. In addition, practical engineering applications are also constrained by strict legal and regulations to ensure safety \citep{RN27}. Thus, before LLMs can be reliably integrated into Hydro-SE, it is essential to rigorously evaluate their domain-specific competence across multidisciplinary knowledge and practical application scenarios.

Benchmarking has become the cornerstone for evaluating the capabilities of LLMs. In general, an LLM benchmark refers to a standardized set of tasks or datasets designed to systematically measure a model’s performance across various dimensions. The concept of benchmarking in natural language processing dates back to early datasets like GLUE \citep{RN29}and SuperGLUE \citep{RN32}(Wang et al., 2019), which were primarily focused on language understanding. With the rapid development of LLMs, a wide range of general-purpose benchmarks—such as MMLU \citep{RN30}, BIG-Bench \citep{RN31}, ARC \citep{RN33}, and GSM8K \citep{RN34} have been proposed to evaluate knowledge, logical reasoning, coding proficiency, and complex problem-solving abilities across different LLMs. These benchmarks provide a common framework for comparison, enabling researchers to quantify progress, identify strengths and weaknesses, and drive further improvements in LLM architectures and training strategies \citep{RN35}.

While these benchmarks provide valuable insights into the general capacities of LLMs, they often fail to capture the specialized expertise required in domain-specific applications. To address this limitation, a growing body of research has focused on vertical-domain benchmarks, which systematically evaluate model performance in specialized fields such as finance \citep{RN16}, law \citep{RN17, RN21}, medicine \citep{RN18}, and chemistry \citep{RN19}. These benchmarks typically include a mix of question formats, including multiple-choice, open-ended reasoning, and applied tasks, to test knowledge recall, professional reasoning, and context-dependent problem solving \citep{RN36}. Despite the rapid development of benchmarks in other scientific domains, a significant research gap remains in the Hydro-SE domain. This field is characterized by highly interdisciplinary knowledge, yet these domain-specific knowledge have received limited representation in both the training and evaluation of most existing LLMs. Therefore, it is valuable to establish a dedicated benchmark for the Hydro-SE domain and systematically evaluate the performance of current LLMs. The benchmark and its evaluation results would serve two main purposes. First, they provide LLM training experts with clear objectives for continued pre-training or supervised fine-tuning \citep{RN37}, enhancing the models’ capabilities in the hydropower domain. Second, they enable researchers in the Hydro-SE domain to understand the strengths and weaknesses of current LLMs, thereby facilitating their effective utilization within the field and supporting more comprehensive and reliable performance in practical applications.

To achieve the above objectives, this study proposes the Hydro-SE Bench, a comprehensive benchmark designed to evaluate LLMs in terms of their Hydro-SE knowledge. Specifically, we develop a multiple-choice benchmark comprising 4,000 questions that span nine core subfields of Hydro-SE. Using this benchmark, we systematically evaluate 10 commercial LLMs and 6 lightweight open-source LLMs, enabling a quantitative comparison of their performance in basic conceptual knowledge, engineering applications, and reasoning ability. The aim of this study is to address the following research questions: (1) To what extent do current general-purpose LLMs possess Hydro-SE specific knowledge and reasoning capabilities? (2) Across which subfields of Hydro-SE do LLMs exhibit stronger or weaker performance? (3) How do the capabilities of LLMs vary across different cognitive levels, specifically, from basic knowledge to practical application and complex calculation in Hydro-SE domain? In addition, this study also estimates the confidence of LLMs when solving the Hydro-SE Bench. Moreover, we provide a mini-subset sampling recommendation for future researchers who aim to evaluate LLM performance in the Hydro-SE domain.

To our knowledge, this is the first Hydro-SE benchmark that integrates multiple subfields and evaluates LLMs’ capabilities in both knowledge and reasoning abilities. The evaluation results provide a foundational reference for understanding the current landscape of LLM competence within the Hydro-SE domain. Building on these findings, Hydro-SE Bench is expected to play an essential role in advancing general-purpose LLMs while also offering actionable insights for developing domain–adapted LLMs and intelligent LLM-based agents. Collectively, these contributions pave the way for more robust and explainable application of LLM in Hydro-SE.

\section{Overview of the Hydro-SE Benchmark}\label{sec2}

The Hydro-SE Bench is presented in Chinese and consists of 4000 question-answer pairs, including 2,700 single-choice questions (SCQ, 68\%) and 1,300 multi-choice questions (MCQ, 32\%). To ensure comprehensive coverage of Hydro-SE-related knowledge, we compiled questions from a wide range of sources (Table \ref{tab:Source materials}), including textbooks, Hydro-SE-related publications, industry standards, and university exams. We adopted a semi-automatic question generation approach (Fig. \ref{fig:Question_generate}) based on prompt-based methods, where expert-designed questions serve as few shot exemplars, and a textual corpus from diverse sources was used as the basis for questions design, thereby expanding the questions pool. For quality assurance, each question was independently reviewed by at least three experts in addition to the original question designer. 

Importantly, the benchmark encompasses nine secondary subfields within the Hydro-SE domain, and all questions are categorized into three distinct classes. The subfields range from general background knowledge to specialized disciplines, such as hydrology and water resources, geotechnical engineering, hydraulic structures, and power system. We categorize all questions into three types (A, B, and C), corresponding respectively to basic conceptual knowledge, scenario-based engineering application, and reasoning and calculations. The distribution of questions across secondary subfields, question types (single and multiple-choice), is presented in Fig. \ref{fig:Question Distri}. Examples of the questions in different subfields and types is detailed in Appendix~\ref{secA1}. Through this hierarchical design, the benchmark aims to comprehensively evaluate LLMs’ multi-dimension capabilities across subfields within the Hydro-SE domain.

\begin{figure}
    \centering
    \includegraphics[trim=1cm 2cm 0cm 3cm, clip, width=1.0\linewidth]{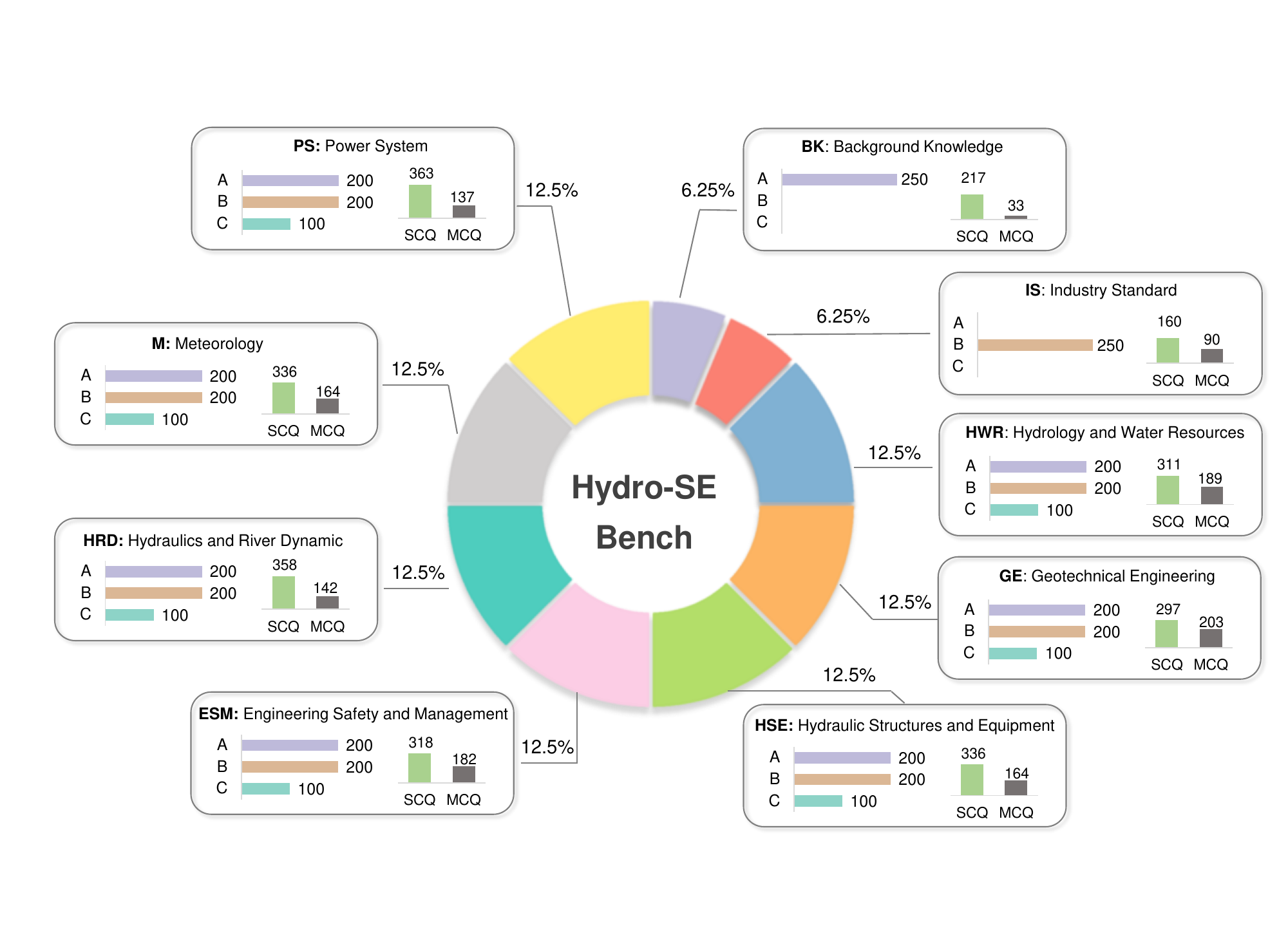}
    \caption{\textbf{Distribution of questions in Hydro-Science and Engineering (Hydro-SE) Benchmark across subfields.} Questions in subfields are manually classified in A (basic conceptual knowledge), B (engineering applications), and C (reasoning and calculation) types. Questions in Hydro-SE Bench comprises single-choice questions (SCQs) and multi-choice questions (MCQs).}
    \label{fig:Question Distri}
\end{figure}

\section{Results}\label{sec3}
\subsection{Overall Performance}\label{subsec1}

To assess the current capabilities of LLMs in the Hydro-SE domain, we evaluated ten commercial models as well as six open-source models with sizes ranging from 7 billion (7B) to 70 billion (70B) parameters using the proposed Hydro-SE Bench. An overview of the LLMs’ performance is presented in Fig. \ref{fig:Overall Perfor}, while the detailed performance statistics are provided in the Appendix~\ref{secA3}. As is shown in Fig. \ref{fig:Overall Perfor}(a), the current state-of-the-art (SOTA) commercial models, with parameter scales exceeding tens of billions, consistently achieve accuracy values in the range of 0.74–0.80 on the Hydro-SE Bench, regardless of their underlying architectures or training origins. This highlights both the impressive general ability of modern LLMs and the remaining gap toward domain-expert-level understanding in Hydro-SE related tasks. Although the overall performance gap among these models remains narrow, DeepSeek-V3.2-Exp and DeepSeek-R1 stand out with the highest overall accuracies, both exceeding 0.79. Meanwhile, Kimi K2 (developed by Moonshot AI) and GLM-4.5 (developed by Zhipu AI) also perform competitively with accuracy above 0.77. Notably, in this round of evaluation, four out of the five top-performing models were developed in China, with GPT-5 (developed by Open AI) ranking fifth. This advantage can be partially attributed to the fact that the current benchmark is provided in Chinese, which may better align with the language priors and fine-tuning corpora of these models. Nevertheless, even considering potential language bias, the consistently strong performance of these models highlights their growing maturity and competitiveness in domain-specific knowledge and reasoning tasks.

Fig. \ref{fig:Overall Perfor}(b) shows the performance of six small-parameter open-source LLMs on the Hydro-SE Bench. The results show that smaller models, with parameter counts in the order of tens of billions or less, achieve accuracy values ranging from 0.41 to 0.68, which is notably lower than those of large-scale commercial LLMs. This observation highlights a clear scaling effect: as models’ parameter size increases, the representational capacity and diversity of learned knowledge also expand, enabling better generalization and higher accuracy on domain-specific tasks. Despite having fewer parameters, GLM-4-32B achieves the best performance among the six models, with an accuracy of 0.68, which is close to that of the Claude-4.5 (Sonnet). The Qwen series ranks second, and the comparison further indicates that parameter scaling from 32B to 72B yields only marginal improvements in accuracy.

Fig. \ref{fig:Overall Perfor} also shows that model performance varies across task A (basic conceptual knowledge), B (engineering applications), and C (reasoning and calculation). Among the ten commercial closed-source models, six achieved the highest accuracy in task A and the lowest in task B. Meanwhile, four models, including DeepSeek-V3.2-Exp, GPT-5, Grok-4, and Gemini-2.5-Pro, performed best on task C, even surpassing their results on task A. For the small-parameter open-source models, most achieved the highest accuracy in task A and the lowest in task C. The only exception was GLM-4-32B-0414, which performed better on task C than on task B. While the superior performance of LLMs on basic conceptual knowledge is expected, the finding that several models achieved higher accuracy in reasoning and calculation than in engineering applications, and sometimes even higher than in knowledge tasks, deviates from our initial assumptions. This result indicates that, despite their strong theoretical and mathematical reasoning capabilities, current LLMs still exhibit considerable limitations when applied to practical Hydro-SE problems. In particular, LLMs often struggle to select appropriate approaches, assess potential impacts, and understand the causes of failures in highly complex and diverse engineering scenarios. Representative examples from the Hydro-SE Bench include generating sediment management strategies for silt-laden reservoirs, summarizing the impacts of urbanization on precipitation, and analyzing the causes of hydraulic turbine failures. These limitations underscore the need for further domain-specific training and enhancement to enable LLMs to reason and make decisions effectively in complex Hydro-SE scenarios.

\begin{figure}
    \centering
    \includegraphics[trim=1cm 0cm 5cm 3cm, clip, width=1\linewidth]{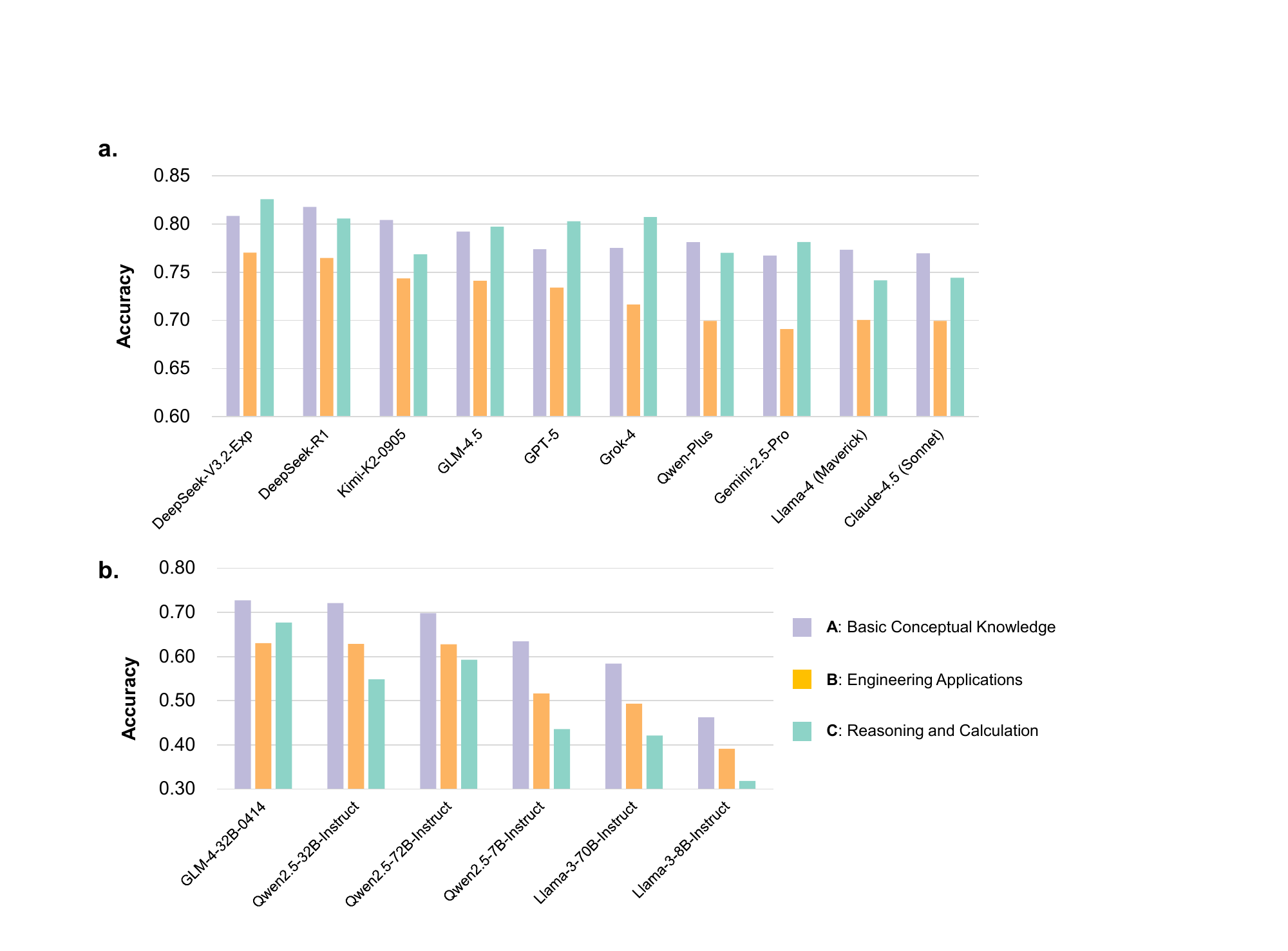}
    \caption{\textbf{Overall performance of LLMs on Hydro-SE Bench.} Accuracy represents the ratio of correctly answered questions to the total number of questions, calculated separately for type A (basic conceptual knowledge), B (engineering applications), and C (reasoning and calculation) questions in the Hydro-SE Bench. (a) The results of ten commercial LLMs. (b) The results of six small-parameter open-source LLMs. In both subfigures, models are arranged from left to right in descending order of overall accuracy.}
    \label{fig:Overall Perfor}
\end{figure}

\subsection{Performance across Different Subfields}\label{subsec2}

\begin{figure}
    \centering
    \includegraphics[trim=4cm 6cm 4cm 5cm, clip, width=1\linewidth]{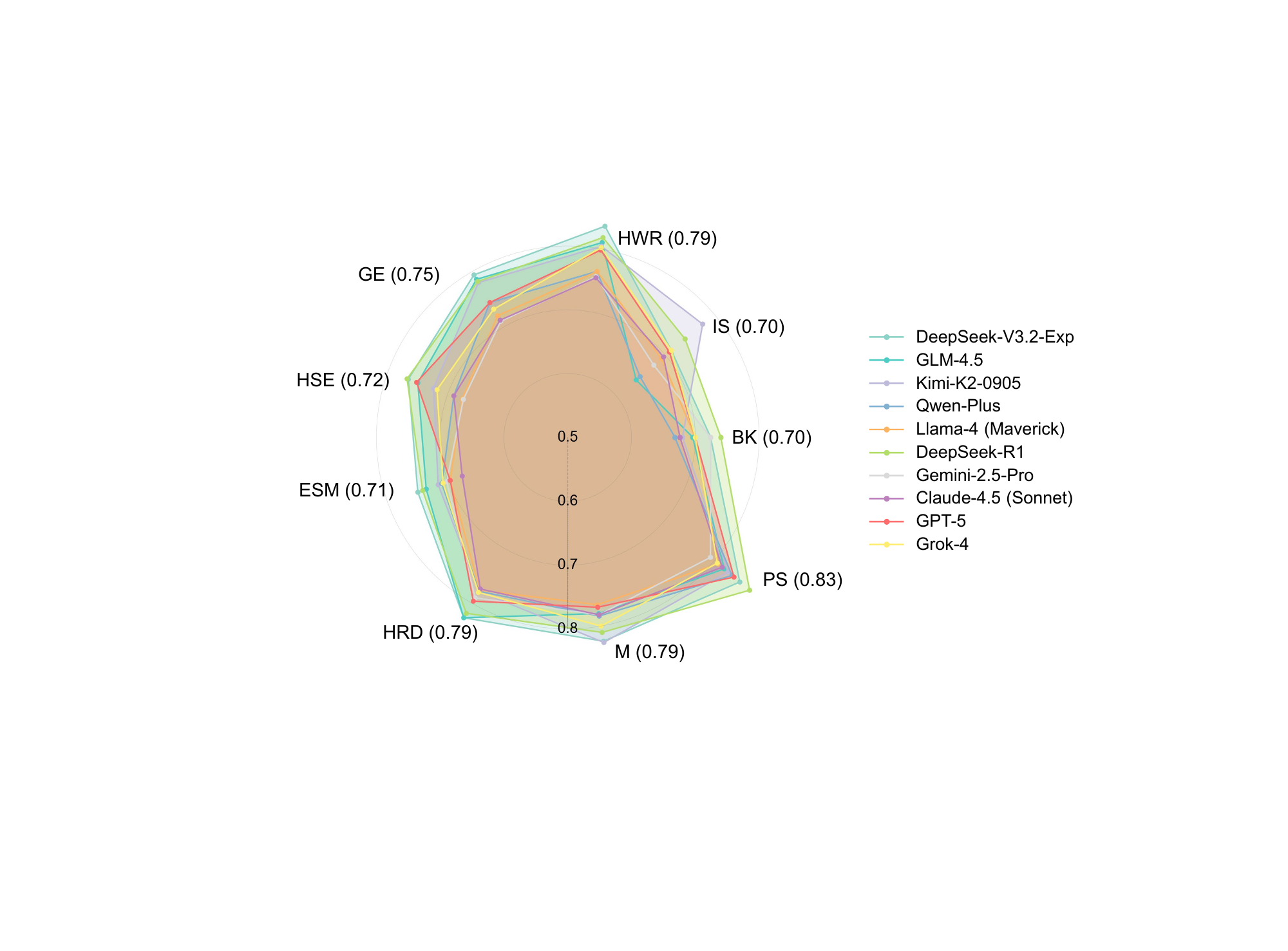}
    \caption{\textbf{Performance of ten commercial LLMs across different subfields of Hydro-SE Bench.} Performance is measured by the ratio of correctly answered questions to the total number of questions, calculated by each subfield. A larger colored area indicates better performance. Subfields are represented by abbreviations (BK: Background Knowledge; IS: Industry Standard; HWR: Hydrology and Water Resources; GE: Geotechnical Engineering; HSE: Hydraulic Structures and Equipment; ESM: Engineering Safety and Management; HRD: Hydraulics and River Dynamic; M: Meteorology; PS: Power System). The numbers in parentheses following each abbreviation denote the average accuracy value of the ten models in that subfield.}
    \label{fig:Subfield Perfor}
\end{figure}

To gain a more detailed understanding of model performance, we manually classified the Hydro-SE Bench into nine subfields and calculated the accuracy for each subfield. Detailed statistical results of different models across all subfields are provided in Fig. \ref{fig:Num_sub_perform}, while the results of ten commercial LLMs are presented in Fig. \ref{fig:Subfield Perfor}. In this radar plot, the minimum accuracy for each dimension is 0 (no question answered correctly) and the maximum accuracy is 1. A larger colored area represents better overall performance.

As illustrated in Fig. \ref{fig:Subfield Perfor}, the performance of LLMs varies across both models and subfields. Subfields such as “Power Systems” (PS), “Meteorology” (M), and “Hydraulics and River Dynamics” (HRD) yield relatively high accuracies, with average values of 0.83, 0.79, and 0.79, respectively. In contrast, “Background Knowledge” (BK), “Industry Standards” (IS), and “Engineering Safety and Management” (ESM) exhibit notably lower accuracy of 0.70, 0.70, and 0.71, respectively. This divergence suggests that LLMs tend to perform better in subfields that are physically grounded and more closely aligned with the natural sciences. Specifically, the subfield of HRD is derived from fluid mechanics and is grounded in well-established physical principles; while M is closely related to the natural sciences, where the underlying mechanisms are relatively universal and less dependent on regional or institutional variations. In contrast, LLMs’ performance declines in domains where knowledge is frequently updated or heavily reliant on specialized terminology. For example, knowledge related to IS and BK is subject to continuous revisions and frequent updates. Meanwhile, the subfield of “Hydraulic Structures and Equipment” (HSE) involves a considerable number of highly specialized terminology and device-specific concepts. Such types of knowledge are not well represented in the static training corpora of current LLMs. Consequently, considerable discrepancies can be observed across different models in these subfields, with some models generating outdated, incomplete, or inaccurate responses.

\subsection{Comparison of LLMs on Different Parameter Scales}\label{subsec3}

While the scaling law has clearly indicated that increasing model size enhances both general and domain-specific capabilities, smaller models still play an important role in practical engineering applications where efficiency and cost are key considerations. Accordingly, this section presents a comparison between large- and small-parameter LLMs in terms of their domain performance, using 10 commercial LLMs evaluated in the study as representatives of large-parameter models and six open-source LLMs as representatives of small-parameter models. 

\begin{figure}
    \centering
    \includegraphics[trim=1cm 7cm 4cm 5cm, clip, width=1\linewidth]{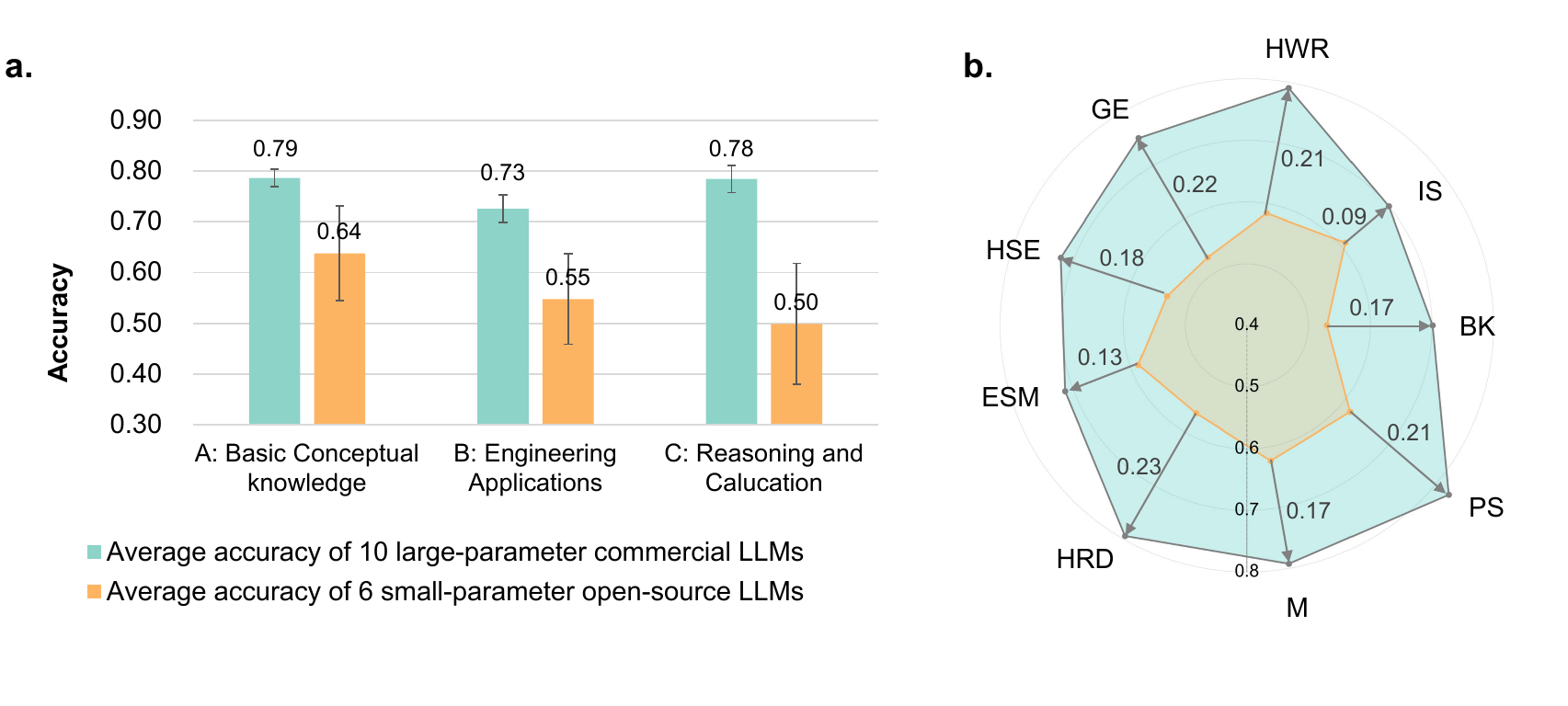}
    \caption{\textbf{Comparison of LLMs across different parameter scales.} (a) The bar chart shows the average accuracy values of large-parameter LLMs and small-parameter LLMs, with error bars representing the standard deviations of model accuracy values. (b) The radar chart illustrates the average accuracy values of large-parameter and small-parameter LLMs across nine Hydro-SE subfields. The numbers marked along the arrows indicate the accuracy differences between large- and small-parameter models.}
    \label{fig:Compare Perfor}
\end{figure}

Fig. \ref{fig:Compare Perfor} (a) shows the average accuracy of 10 evaluated commercial LLMs and 6 small-parameter open-source LLMs, with error bars representing the standard deviations of model accuracy values. With the parameters increasing, the most significant improvement is observed in task C, where the average accuracy value rises from 0.50 to 0.78, representing an increase of 56.0 \%. Meanwhile, the improvements in task A and task B are relatively modest, with the accuracy value increases of 23.4 \% and 32.7 \%, respectively. These results, building on established scaling laws, further reveal that in the Hydro-SE domain, increasing model parameters primarily enhances reasoning and calculation capabilities. Without incorporating domain-specific knowledge into the training corpus, or performing dedicated continuous pre-training and fine-tuning within specialized Hydro-SE contexts, model performance in conceptual knowledge and engineering applications tasks remains limited.

Fig. \ref{fig:Compare Perfor} (b) compares the average accuracy of large-parameter and small-parameter LLMs across nine Hydro-SE subfields. As shown in Fig. \ref{fig:Compare Perfor} (b), increasing model size yields the most substantial performance gains in “Hydraulics and River Dynamics” (HRD) and “Geotechnical Engineering” (GE), where the absolute accuracy improves by 0.23 and 0.22, respectively, corresponding to relative improvements exceeding 40 \%. In contrast, the smallest gains appear in “Industry Standards” (IS) and “Engineering Safety and Management” (ESM), with absolute improvements of only 0.09 and 0.13, representing 15 \% and 21 \% relative increases, respectively. This pattern aligns with the conclusions drawn from Fig. \ref{fig:Compare Perfor} (a). Tasks in HRD and GE generally contain more reasoning-intensive elements, such as multi-step physical inference, numerical approximation, and stability judgment, where performance tends to scale more directly with parameter count. By comparison, IS and ESM rely heavily on the recall of codified rules, statutory constraints, and procedural criteria. Without corresponding expansion of domain-specific materials in the training corpus, merely increasing parameter size offers limited benefit, resulting in smaller accuracy gains. 

These findings further highlight the key implication that scaling alone cannot compensate for insufficient domain coverage. While larger models improve general reasoning and calculation capabilities, achieving strong performance in Hydro-SE-specific conceptual understanding and engineering applications ultimately requires targeted domain-adaptive pre-training, curated knowledge integration, or additional supervised fine-tuning.

\subsection{Confidence Estimation}\label{subsec4}

LLM confidence reflects the model’s self-evaluated likelihood of the generated output being factually correct. Given that hallucination in LLMs constitutes a major obstacle in real-world application, quantifying output confidence is essential for reliable model evaluation (L Huang et al., 2025). In this study, two top-performing models at different parameter scales (DeepSeek-V3.2-Exp and GLM-4-32B) were prompted to investigate how LLMs can correctly estimate the accuracy of their own responses in the Hydro-SE Bench. For each model, we asked it to output the confidence level for its own answer and used these results to draw the calibration curves.

\begin{figure}[htbp]
    \centering
    \begin{subfigure}[b]{0.48\textwidth}
        \centering
        \includegraphics[width=\textwidth]{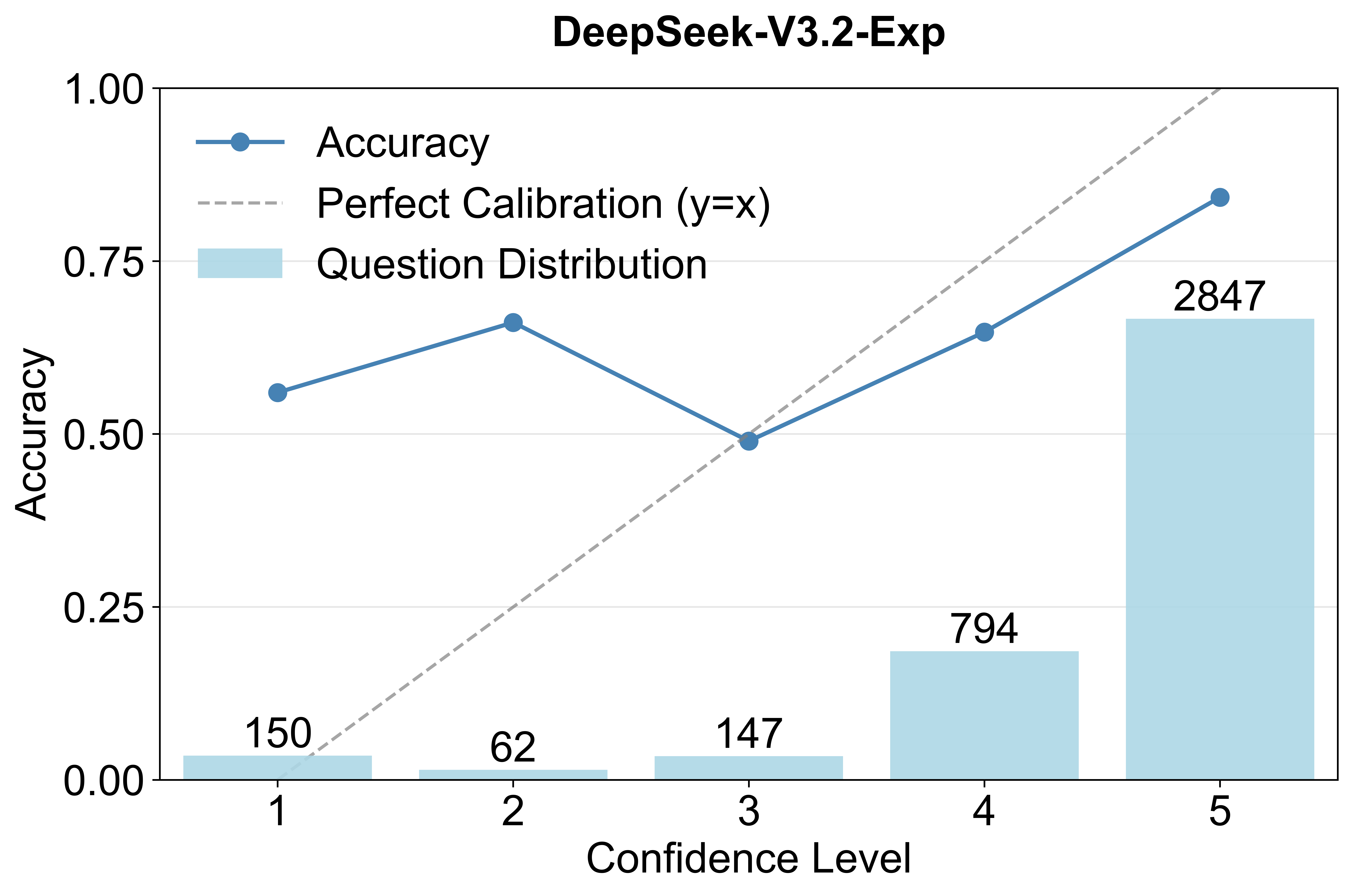}
        \label{fig:Deepseek_confidence}
    \end{subfigure}
    \hfill 
    \begin{subfigure}[b]{0.48\textwidth}
        \centering
        \includegraphics[width=\textwidth]{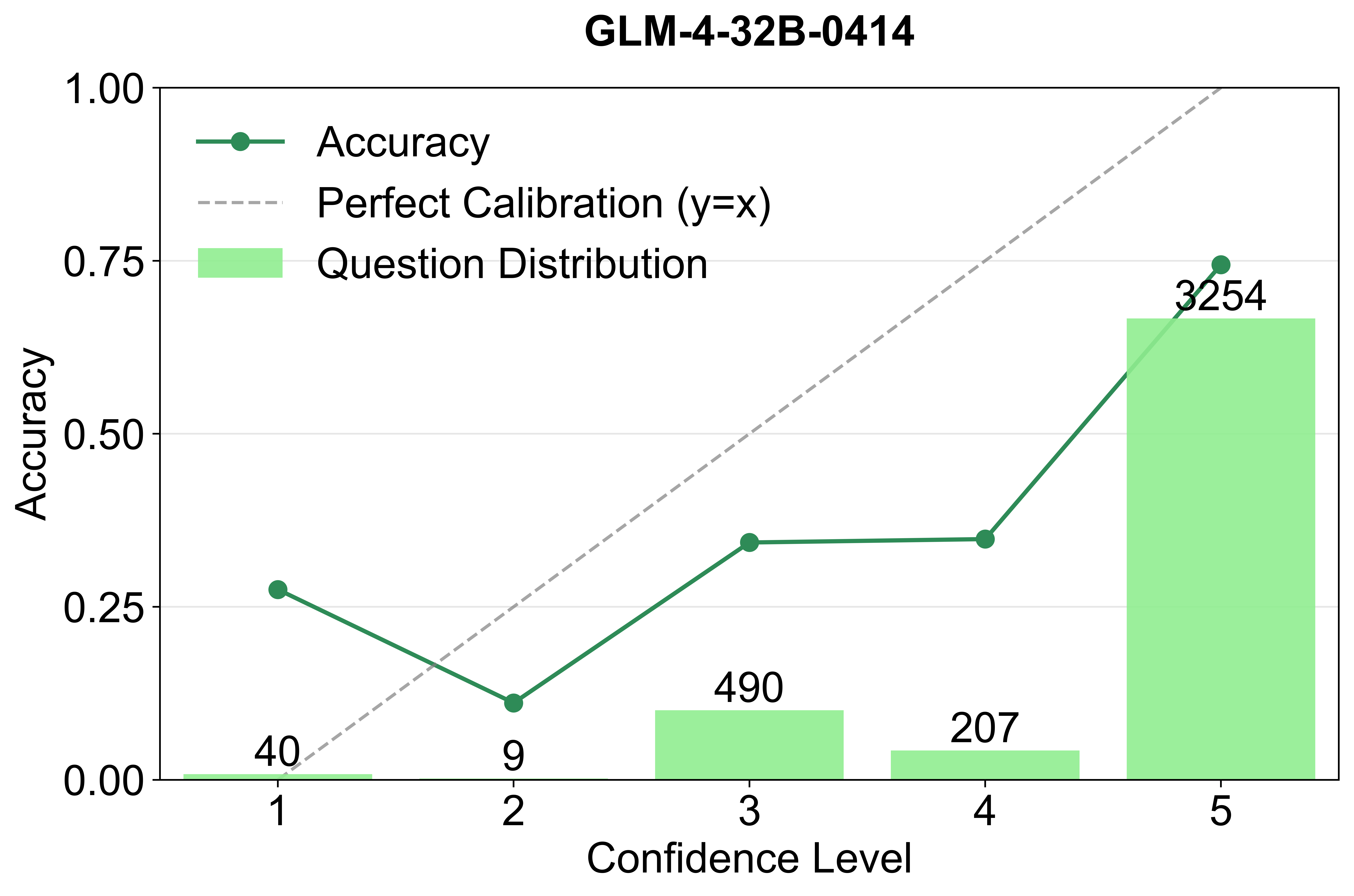}
        \label{fig:GLM_confidence}
    \end{subfigure}
    
    \caption{\textbf{Distribution of LLMs’ confidence estimates for Hydro-SE Bench.} The bar plot shows the distribution of LLMs’ confidence estimates (with the number of questions labeled by the height of the bar). The line plot shows the average accuracy for questions in each confidence level. LLMs’ confidence estimate would be well calibrated if the average accuracy increases with the confidence. The dash black line represents the prefect calibration line, where accuracy increases monotonically with the confidence level.}
    \label{fig:Confidence_estimation}
\end{figure}

As shown in Fig. \ref{fig:Confidence_estimation}, both DeepSeek-V3.2-Exp and GLM-4-32B tend to assign the highest confidence score to the majority of their own answers. Specifically, GLM rated 81.4 \% questions with a confidence level of 5, while DeepSeek did so for 71.2 \% questions. For LLMs, evaluating the correctness of their own sampled responses remains a highly challenging task. In Fig. \ref{fig:Confidence_estimation}, the calibration curve of GLM-4-32B lies consistently below the perfect calibration line, indicating that the model tends to overestimate the correctness of its answers in most cases. For instance, among the questions where GLM assigned a confidence score of 5, its actual accuracy was 74.4 \%, whereas for those rated with a confidence score of 4, the actual accuracy dropped sharply to only 34.8 \%. This mismatch between subjective confidence and objective correctness highlights a notable miscalibration problem, which can be risky in real-world engineering scenarios where misplaced confidence may lead to misleading or unsafe recommendations.

Compared with GLM-4-32B, DeepSeek-V3.2-Exp shows a substantial improvement in calibration performance, which can be partly attributed to its larger parameter scale. When the model assigns confidence levels of 3, 4, or 5, its actual accuracy increases monotonically with the confidence level, and the calibration curve closely aligns with the perfect calibration line, although a slight tendency to overestimate can still be observed. This suggests that DeepSeek exhibits a more reliable mapping between confidence and correctness in the medium-to-high confidence range. Nonetheless, for lower confidence scores, DeepSeek tends to be underconfident, as its calibration curve lies above the perfect line. In this region, the model underestimates its actual accuracy, indicating the presence of asymmetric calibration depending on the confidence level. On average, DeepSeek returns an average confidence score of 4.64 for all the correct answers and 4.15 for incorrect ones. These findings emphasize that calibration quality varies across models and confidence ranges. When deploying LLMs for Hydro-SE applications, calibration should also be treated as a first-class evaluation dimension, alongside domain knowledge and reasoning accuracy.

\section{Methods}\label{sec4}
\subsection{Question generated workflow}\label{subsec5}

To construct Hydro-SE Bench, we utilized a wide board of sources, ranging from textbooks, industry standards to statistic year books Table \ref{tab:Source materials}. We first collect questions from textbook exercises and further invite domain experts to manually design additional ones. In total, 1,592 questions were compiled and carefully annotated by experts with their corresponding subfields and categories (A, B, or C). Each question is accompanied by a detailed explanation, and all answers can be verified against authoritative sources such as published books and official industry standards.

\begin{table}[h]
\caption{\textbf{Source materials for questions generation.} The table shows the category and the number of items as well as a brief description.}\label{tab:Source materials}
\begin{tabular}{@{}lcc@{}}
\toprule
\textbf{Category} & \textbf{Number of Items} & \textbf{Description} \\
\midrule
Books & 985 & \makecell{Hydro-SE related books \\ (including textbooks, reference books, monographs)}\\
\addlinespace
Industry Standards & 600 & \makecell{Chinese national standards \\ and other industry standards} \\
\addlinespace
Laws and Regulations & 57 & \makecell{Water management related laws \\ and regulations} \\
\addlinespace
Statistical Yearbooks & 1113 & \makecell{Bulletins on multi-level water resources,\\ soil water management, etc.}\\
\bottomrule
\end{tabular}
\end{table}

\begin{figure}
    \centering
    \includegraphics[trim=0cm 3cm 2cm 4cm, clip, width=1\linewidth]{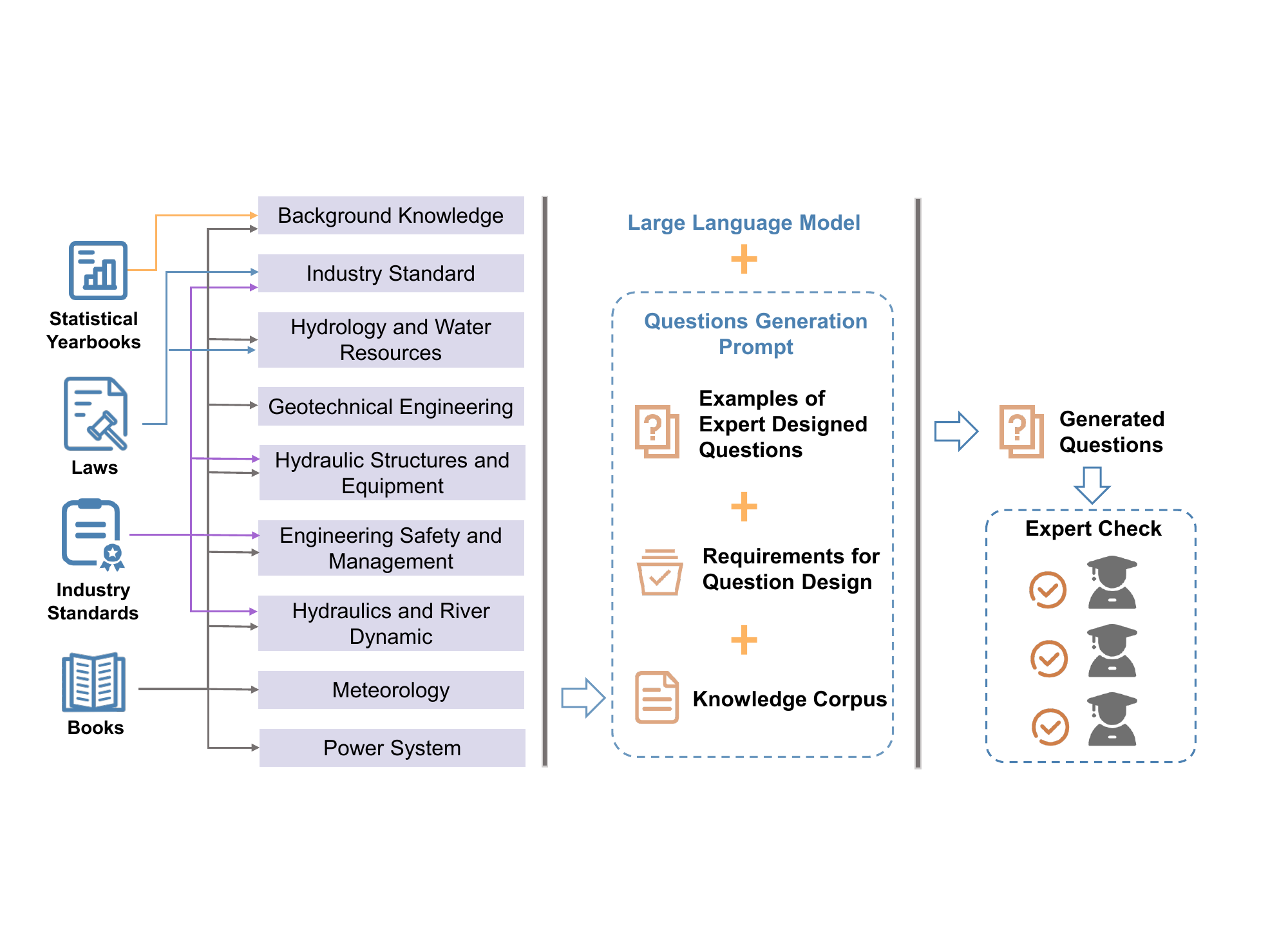}
    \caption{\textbf{Semi-automatic question generation approach.} }
    \label{fig:Question_generate}
\end{figure}

Based on these expert-designed questions, we developed a semi-automatic question generation approach as shown in Fig. \ref{fig:Question_generate}. We first categorized the source materials according to nine distinct subfields in the Hydro-SE Bench. The original documents were then parsed and segmented into smaller passages, which were subsequently provided to LLMs as contextual inputs. Using prompts (see Table \ref{tab:question-generate-prompt}) and expert-designed questions as the few-shot, we guided the models to generate questions directly based on the content of each knowledge slice. This approach enables convenient future expansion of the benchmark, as new documents can be parsed and added following the same pipeline. In addition, the generated questions have clearly traceable references, ensuring the transparency and verifiability of each answer. To guarantee the quality of LLM-generated questions, we arranged at least three rounds of expert review and refinement to these questions.

\subsection{Model evaluation workflow}\label{subsec6}
We adopted consistent prompt templates across all evaluated LLMs (see Table \ref{tab: LLM eval prompt}) to ensure the fairness and comparability of the tests. In the first stage, each model was prompted to generate its complete reasoning (if applicable) or problem-solving process. Subsequently, the response context was provided to another LLM, which was then prompted to extract the final choice letters from the original response. This two-stage approach was designed to capture the models’ genuine reasoning processes rather than merely constraining them to output choice letters in specific format. Its main advantage is that complete reasoning chains reduce interpretability and provide a more faithful evaluation of the models’ capabilities \citep{RN14}. In addition, this design avoids the constraints of strict instruction-following, which could otherwise bias the evaluation of their intrinsic reasoning and Hydro-SE domain knowledge \citep{RN38}. 

For all models, responses were sampled with the temperature set to 0 to ensure deterministic outputs. We accessed the models via their official API endpoints or through the OpenRouter platform. DeepSeek-V3.2-Exp was employed to extract choice letters from the responses, given its strong performance in text understanding. Moreover, since our primary goal was to evaluate the models’ knowledge capabilities rather than their robustness to interface issues, we re-sampled the responses whenever an API call failed or the output was truncated, instead of marking those cases as incorrect answers.

\subsection{Confidence estimation method}\label{subsec7}
To estimate the models’ confidence, we employed a verbalized confidence estimation approach \citep{RN40}. Specifically, LLMs were prompted with both the question (including the answer options for multiple-choice questions) and their own selected answer, and then asked to rate their confidence in the proposed answer on a scale from 1 to 5. The detailed prompt is provided in Table \ref{tab: confidence estimation prompt}. This method was chosen because most closed-source models, such as GPT-5, do not provide access to logits, making the verbalized approach more generally applicable. Moreover, this approach is considered closer to practical use cases and enables capturing semantic uncertainty \citep{RN19}, which is distinct from the probability of a token given a sequence of tokens. Based on the outputs of LLMs, the questions in Hydro-SE Bench were first divided into five subsets according to the confidence levels reported by the models. For each subset, we then calculated the models’ accuracy and plotted calibration curves. When higher predicted confidence aligns with higher actual accuracy, it suggests that the models’ self-evaluation is well-calibrated and they are less prone to hallucinations.

\section{Discussion}\label{sec5}
\subsection{Difficulty Consistency Analysis}\label{subsec8}

In this study, the constructed Hydro-SE Bench comprises a total of 4,000 questions. Completing the entire benchmark requires a total of 83.9 million tokens for SOTA models and 15.3 million tokens for open-source lightweight models (see Table \ref{tab:llm_token}), leading to considerable computational and time costs. To enable more efficient routine evaluations, we conducted a sampling analysis on Hydro-SE Bench to explore the feasibility of constructing smaller benchmark subsets. This analysis examined the difficulty consistency among questions within types A, B, and C, and provided a reference for determining the minimum recommended sampling ratio. The sampling procedure was designed to maintain representativeness across all subfields. Specifically, for each subfield, questions of types A, B, and C were sampled at the same proportional rate. The sampled questions from all subfields were then aggregated to form a subset, on which model accuracy was evaluated. Sampling ratios ranged from 10 \% to 100 \% in increments of 10 \%, resulting in ten sampling ratio levels. For each level, ten independent random samplings were conducted, and the mean and standard deviation of the accuracy were calculated across these repetitions. In this study, DeepSeek-V3.2-Exp and Claude-4.5, representing the highest- and lowest-scoring models among the commercial models evaluated, are used as examples to illustrate how model performance varies across subsets with different sampling ratios of the full benchmark.

\begin{figure}[htbp]
    \centering
    \begin{subfigure}[b]{0.48\textwidth}
        \centering
        \includegraphics[width=\textwidth]{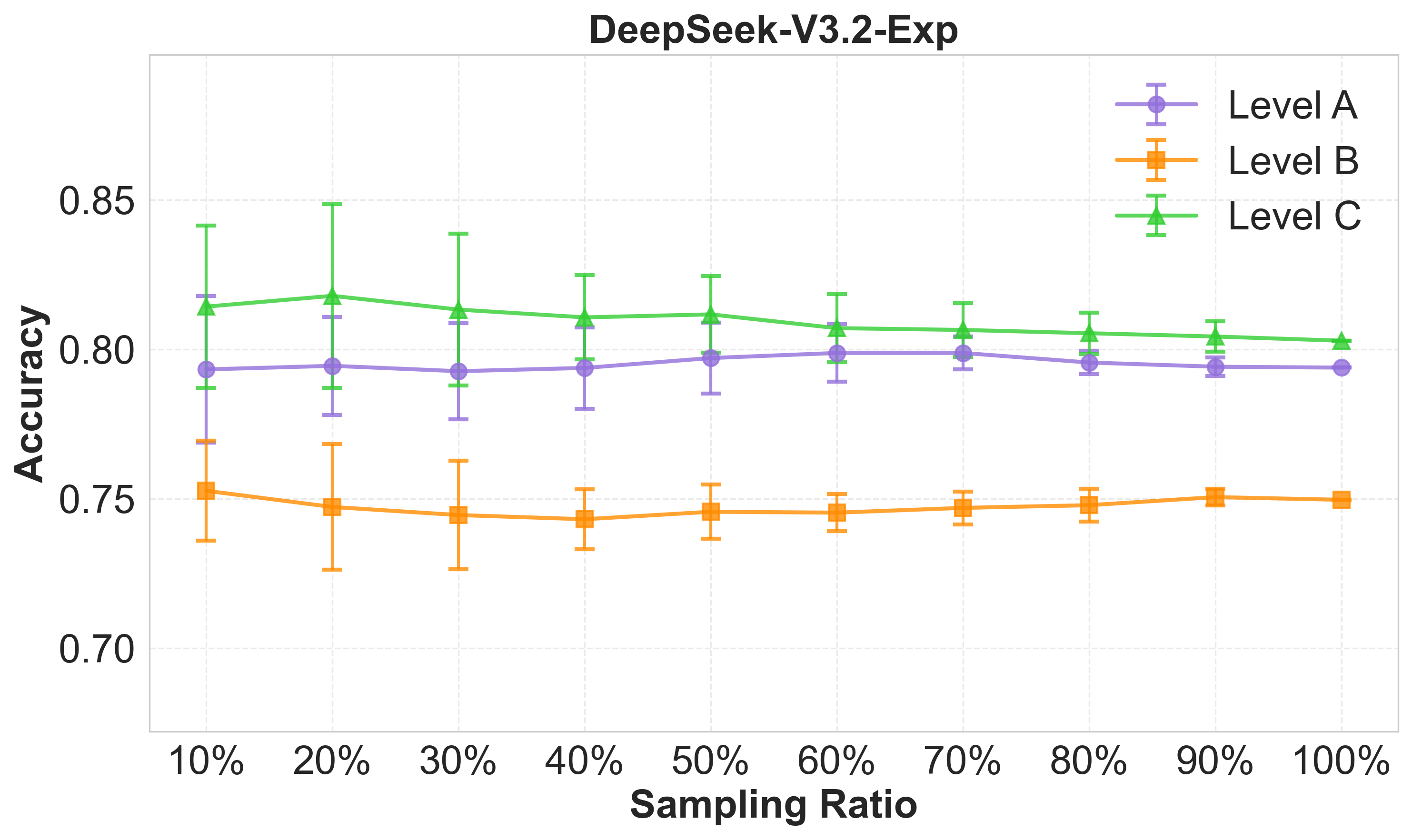}
        \label{fig:Deepseek_sample}
    \end{subfigure}
    \hfill 
    \begin{subfigure}[b]{0.48\textwidth}
        \centering
        \includegraphics[width=\textwidth]{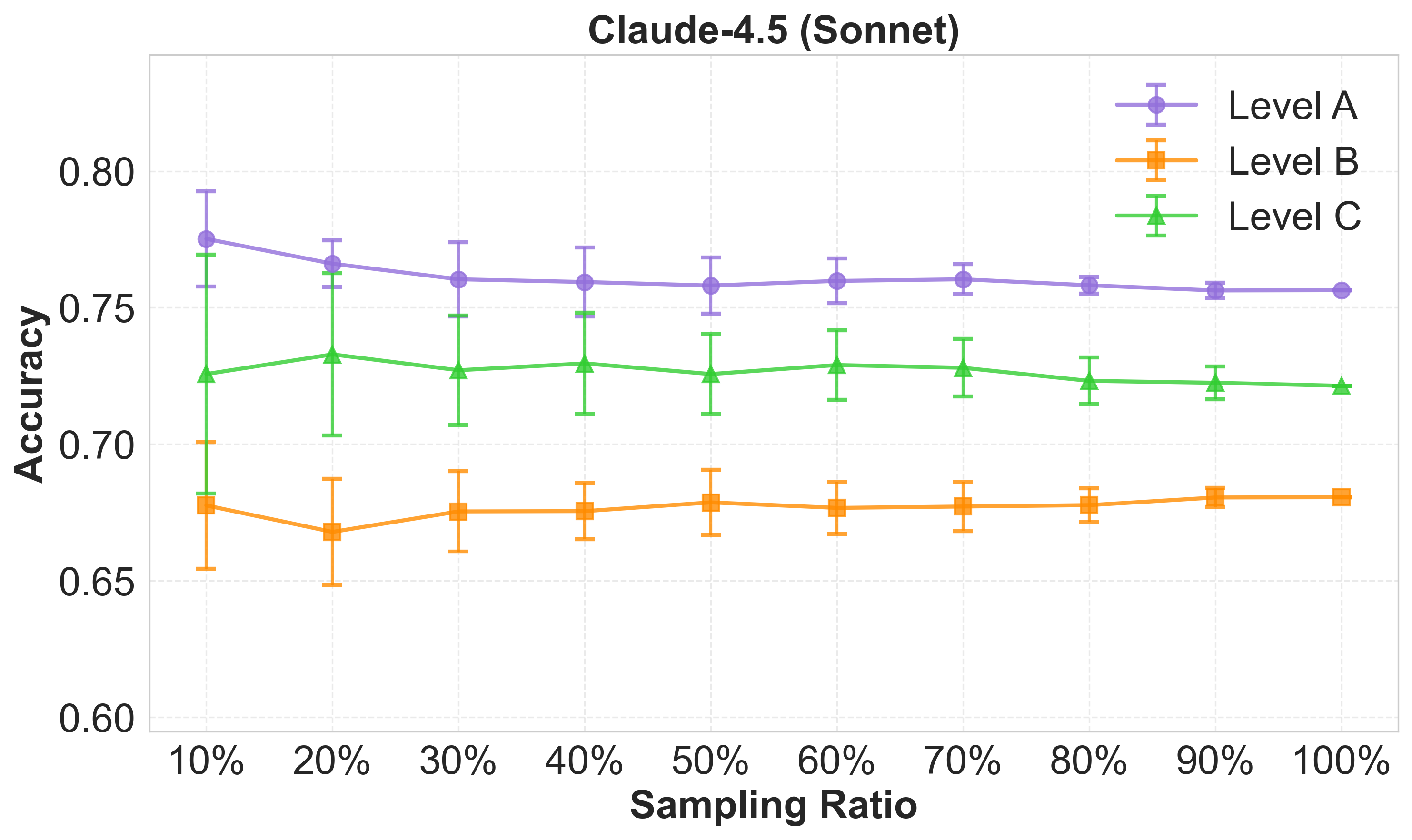}
        \label{fig:Claude_sample}
    \end{subfigure}
    
    \caption{\textbf{LLM performance across subsets with different sampling ratios of the full Hydro-SE Bench.} Line plots represent model accuracy on question types A (basic concepts), B (engineering applications), and C (reasoning and calculations). Error bars indicate the standard deviations obtained from ten random sampling repetitions at each sampling ratio.}
    \label{fig:Smapling_accuracy}
\end{figure}

Fig. \ref{fig:Smapling_accuracy} shows the sampling evaluation results of DeepSeek-V3.2-Exp and Claude-4.5. The line plots show the model's accuracy on sampled subsets, with error bars indicating the standard deviation of multiple samplings. As shown in Fig. \ref{fig:Smapling_accuracy}, the accuracy of both DeepSeek-V3.2-Exp and Claude-4.5 (Sonnet) remain relatively stable across different sampling ratios of Hydro-SE Bench, indicating good difficulty consistency within the benchmark. Across all sampling ratios from 10 \% to 100 \%, the difference between the maximum and minimum average accuracy is less than 1.9 \% for DeepSeek-V3.2-Exp and 1.5 \% for Claude-4.5 (Sonnet). When the sampling ratio $\geq$ 30 \%, the total range of the error bars does not exceed 5 \%, which indicates that random sampling introduces relatively minor fluctuations. However, when the sampling ratio drops below 30 \%, the accuracy variation between different samplings becomes more pronounced, particularly for type C questions, as reflected by longer error bars. Based on these results, it is recommended that when constructing subsets of Hydro-SE Bench for testing, the sampling ratio for each question type within subfields should be at least 30 \%.

\subsection{Limitations and future work}\label{subsec9}

While this study provides systematic evaluation of LLMs’ Hydro-SE-related capabilities, several limitations remain. First, Hydro-SE Bench focuses primarily on multiple-choice questions, which capture a wide range of knowledge and reasoning abilities but still cannot fully reflect the open-ended and dynamic decision-making processes required in real engineering applications. Second, although the benchmark covers nine representative subfields, it does not yet incorporate real-time data or scenario-based agentic tasks that would more closely resemble practical scenarios. Third, the evaluation primarily centers on text-based reasoning, whereas many Hydro-SE applications rely on multimodal information such as satellite remote sensing imagery, time-series data collected by stations, and video streams captured by surveillance cameras, which future benchmarks should integrate.

Future research could further expand Hydro-SE Bench to include interactive, multimodal, and scenario-driven tasks, enabling the evaluation of LLMs both as knowledge bases and decision-support agents. In addition, establishing standardized metrics for domain reasoning, uncertainty estimation, and human–AI collaboration will also be crucial for advancing the safe and reliable applications of LLMs in Hydro-SE. Looking forward, we believe that continuous benchmarking and longitudinal tracking of model updates will be essential for revealing performance trajectories, diagnosing persistent weaknesses, and guiding targeted fine-tuning efforts. Such practices are crucial for transforming current LLMs from general-purpose tools into reliable assistants for Hydro-SE.

\section{Conclusion}\label{sec6}
In this study, we introduce Hydro-SE Bench, the first comprehensive benchmark designed to evaluate LLMs’ knowledge and application abilities in the Hydro-SE domain. Through evaluating sixteen LLMs (10 commercial LLMs and 6 small-parameter open source LLMs) across nine Hydro-SE subfields and three task categories, we delineate both the strengths and limitations of current LLMs and provide a quantitative comparison between models across different parameter scales. The evaluation results on Hydro-SE Bench indicate that overall accuracy ranges from 0.74–0.80 for large-parameter commercial LLMs and 0.41–0.68 for small-parameter open-source models. While LLMs demonstrate strong performance in subfields grounded in natural and physical sciences, they continue to struggle with knowledge areas characterized by frequent updates, specialized terminology, or engineering-specific conventions. Parameter scaling primarily enhances models’ reasoning and calculation abilities, with an average improvement of 56.0 \%, yet brings far more limited gains in basic conceptual knowledge and engineering application. These findings highlight that while current LLMs can serve as valuable assistants for Hydro-SE applications, they are not yet reliable for practical applications. Moreover, scaling alone cannot compensate for insufficient domain-specific training. Targeted domain pre-training and supervise fine-tuning remains essential. In addition, this study examined the calibration behavior of LLMs and assessed the difficulty consistency of Hydro-SE Bench to support efficient sampling-based evaluation.

The evaluation framework established in this study provides both a baseline and a roadmap for improving domain-specific LLMs. By revealing LLMs’ performance gaps between conceptual knowledge, engineering applications, and reasoning and calculation, Hydro-SE Bench offers clear guidance for model developers seeking to enhance domain alignment and for Hydro-SE researchers exploring the integration of LLMs into real-world decision processes. Ultimately, Hydro-SE Bench represents an early but important step toward building reliable, domain-specialized LLMs capable of supporting sustainable and intelligent Hydro-SE management.

\vspace{1cm}

\bmhead{Supplementary information}
Appendix \ref{secA1}: Example questions in Hydro-SE Bench;
Appendix \ref{secA2}: Prompts applied in this study;
Appendix \ref{secA3}: Model performance statistics;\\

\bmhead{Acknowledgements}
This study was supported by the National Natural Science Foundation of China (Grant No. 52539001).\\

\phantomsection
\bmhead{Competing interests}
The authors declare no competing interests.\\

\bmhead{Data availiability}

The Hydro-SE Bench and implementation code is released under the MIT license and can be accessed at: https://github.com/sheishijun/Hydro-SE-Bench

\bibliography{sn-bibliography}


\newpage
\begin{appendices}

\section{Example questions in Hydro-SE Bench}\label{secA1}
\small
\begin{longtable}{L{0.3\textwidth} p{0.7\textwidth}}
\caption{Basic Conceptual Knowledge Questions (Type A)}\label{tab: basic_concept} \\
\toprule
\textbf{Subfield} & \textbf{Question and Options} \\
\midrule
\endfirsthead 

\multicolumn{2}{c}%
{{\bfseries \tablename\ \thetable{} -- continued from previous page}} \\
\toprule
\textbf{Subfield} & \textbf{Question and Options} \\
\midrule
\endhead 

Background Knowledge & Which of the following is the largest freshwater lake in China? \newline A. Qinghai Lake; B. Dongting Lake; C. Poyang Lake; D. Tai Lake. \\
\addlinespace
Hydrology and Water Resources & During the hydrological cycle, which process serves as the primary linkage for the interaction between surface water and groundwater, and its rate is significantly influenced by factors such as soil types, vegetation covers, and rainfall characteristics? \newline A. Evaporation; B. Infiltration; C. Runoff; D. Precipitation. \\
\addlinespace
Geotechnical Engineering & What is the primary function of a settlement joint? \newline A. To prevent differential settlement of the foundation; \newline B. To enhance the overall stability of the foundation; \newline C. To improve the load-bearing capacity of the foundation; \newline D. All of the above. \\
\addlinespace
Hydraulic Structure and Equipment & When a hydropower station is characterized by high water head, low flow, and a long penstock, which of the following water conveyance schemes is typically adopted? \newline A. Unit conveyance; B. Combined conveyance; \newline C. Group conveyance; D. All of the above. \\
\addlinespace
Engineering Safety and Management & In China, after completing a water conservancy construction project, which of the following is not included in the content of the final acceptance? \newline A. Project quality; \newline B. Record of project schedule implementation; \newline C. Project investment; \newline D. Environmental protection measures. \\
\addlinespace
Hydraulics and River Dynamic & An ideal fluid is defined as a fluid that ( ) .\newline A. neglects gravitational effects; \newline B. exhibits uniform internal pressure under static conditions; \newline C. experiences no viscous resistance during flow; \newline D. maintains uniform internal pressure during flow. \\
\addlinespace
Meteorology & An increase in atmospheric moisture content due to the evaporation of surface water will cause the dew point temperature to ( ). \newline A. rise; B. fall; C. remain unchanged; D. be indeterminable. \\
\addlinespace
Power System & In the Chinese power system, how is the grounding of the neutral point in the primary winding of a voltage transformer typically classified? \newline A. Working grounding; B. Protective grounding \newline C. Safety grounding; D. Neutral grounding connection. \\
\bottomrule
\end{longtable}

\small
\begin{longtable}{L{0.3\textwidth} p{0.7\textwidth}}
\caption{Engineering Application Questions (Type B) }\label{tab:engineering_app} \\
\toprule
\textbf{Subfield} & \textbf{Question and Options} \\
\midrule
\endfirsthead 

\multicolumn{2}{c}%
{{\bfseries \tablename\ \thetable{} -- continued from previous page}} \\
\toprule
\textbf{Subfield} & \textbf{Question and Options} \\
\midrule
\endhead 

\endfoot 

\bottomrule 
\endlastfoot 

Industry Standard & According to the Flood Control Law of the People's Republic of China, which of the following is the correct classification of flood control zones? \newline A. Flood basins, flood storage and detention basins, and areas protected against floods; \newline B. Waterlogging-prone areas, drought-prone areas, and saline-alkaline areas; \newline C. Mountain torrent area, plain areas, and coastal areas; \newline D. Major protected areas and general protected areas. \\
\addlinespace
Hydrology and Water Resources & In urban flood forecasting, which of the following models is generally not adopted? \newline A. SWMM model; B. SCS model; \newline C. MIKEURBAN model; D. Xinanjiang model. \\
\addlinespace
Geotechnical Engineering & During the construction of a filter layer for an earth-rock dam, which of the following represents a correct quality control requirement? \newline A. One sample shall be taken for every 100 m² of filled area; \newline B. For a strip-shaped filter layer, one sampling cross-section shall be arranged every 20 meters; \newline C. Verify whether the filter coefficient $D_{50}/d_{50}$ complies with the design requirements; \newline D. Allow minor amounts of extraneous matter that do not affect performance; \newline E. The deviation in placement thickness shall be controlled within ±5 cm. \\
\addlinespace
Hydraulic Structure and Equipment & During the hydraulic design of a metal spiral case, what is the key step that must be completed first? \newline A. Determine the connection structure between the spiral case and the stay ring; \newline B. Select the wrap angle $\Phi_0$ and the average flow velocity $\bar{v}_0$ at the inlet section; \newline C. Calculate the hydraulic coefficient $C$ of the spiral case; \newline D. Draw the single-line plan of the spiral case; \newline E. Determine the specific dimensions of the butterfly transition section of the stay ring. \\
\addlinespace
Engineering Safety and Management & In China, which of the following nondestructive testing techniques is commonly used for quality inspection of internal concrete defects in sluice structures? \newline A. Ground-penetrating radar detection technology; \newline B. Core drilling sampling method; \newline C. Ultrasonic testing technology; \newline D. All of the above. \\
\addlinespace
Hydraulics and River Dynamic & A city in the middle reaches of the Yangtze River plans to renovate the flood control levee of its riverside park. The levee structure must meet two requirements: it should have the capacity to withstand a 50-year flood event, while also creating a suitable ecological environment for fish spawning. Among the following options, which type of levee design best fulfills both requirements? \newline A. Vertical reinforced concrete floodwall; \newline B. Gently-sloped ecological bag revetment; \newline C. Stepped-fishway composite ecological levee; \newline D. Elevated pile-supported viewing platform levee. \\
\addlinespace
Meteorology & During August, when the Western Pacific Subtropical High Ridge Line stably maintained between 30°N and 35°N, which of the following typical weather phenomena is most likely to occur in the region? \newline A. Jiang -Huai Meiyu (Plum Rains); \newline B. North China rainy season; \newline C. Autumn crisp weather in South China; \newline D. Western China autumn rains. \\
\addlinespace
Power System & In the Chinese power system, which category of equipment has its insulation level primarily determined by lightning overvoltage? \newline A. 220 kV substation equipment; \newline B. Substation equipment rated 330 kV and above; \newline C. Ultra-high voltage equipment after switching overvoltage limitation; \newline D. Valve groups in DC converter stations. \\
\bottomrule
\end{longtable}

\small
\begin{longtable}{L{0.3\textwidth} p{0.7\textwidth}}
\caption{Reasoning and Calculation Questions (Type C) }\label{tab:reasoning_calcu} \\
\toprule
\textbf{Subfield} & \textbf{Question and Options} \\
\midrule
\endfirsthead 

\multicolumn{2}{c}%
{{\bfseries \tablename\ \thetable{} -- continued from previous page}} \\
\toprule
\textbf{Subfield} & \textbf{Question and Options} \\
\midrule
\endhead 

\endfoot 

\bottomrule 
\endlastfoot 

Hydrology and Water Resources & For a basin area of 2916 km², the sum of the ordinates of its 6-hour unit hydrograph is known to be 11475 m³/s. What is the runoff depth corresponding to this unit hydrograph? \newline A. 10 mm; B. 85 mm; C. 22.25 mm; D. 15 mm; E. 50 mm. \\
\addlinespace
Geotechnical Engineering & For a foundation pit requiring a dewatering depth of 6 m, the thickness of the unconfined aquifer is 15 m, the permeability coefficient k = 15 m/d, the radius of influence R = 150 m and the equivalent radius $r_0$ = 30 m. According to the calculation for a fully penetrating well in an unconfined aquifer, which of the following options is closest to the total water inflow of the foundation pit? \newline A. 2240 m³/d; B. 2580 m³/d; C. 3024 m³/d; D. 3787 m³/d. \\
\addlinespace
Hydraulic Structure and Equipment & For a Kaplan turbine operating under a design head of 18.6 m with a discharge of 825 m³/s, the turbine efficiency is 0.86 and the generator efficiency is 0.968. Neglecting the influence of hydraulic losses, which of the following options provide correct calculation results? \newline A. Water power output is 150,500 kW; \newline B. Turbine output is 129,430 kW; \newline C. Unit output is 125,288 kW; \newline D. Overall efficiency is 83.2\%; \newline E: Input power to the turbine is 150,500 kW. \\
\addlinespace
Engineering Safety and Management & The supervision contract for a certain project stipulates that the supervision service fee shall be calculated as 3.5\% of the project budget estimate, with the design phase accounting for 30\%, the construction phase for 60\%, and the acceptance phase for 10\%. Given that the total project budget estimate is 280 million yuan and the actual investment during the construction phase is 160 million yuan, which of the following calculation results is correct? \newline A. The supervision fee for the design phase is 29.4 million yuan; \newline B. The supervision fee for the construction phase is 168 million yuan; \newline C. The supervision fee for the acceptance phase is calculated as 10\% of the actual cost during the construction phase; \newline D. The total supervision contract amount is 9.8 million yuan. \\
\addlinespace
Hydraulics and River Dynamic & In the calculation of the water surface profile for a natural river channel, given a reach length $\Delta s$ = 800 m, downstream water level $z_d$ = 85.3 m, cross-sectional area $A_d$ = 120 m², wetted perimeter $\chi_d$ = 45 m, and discharge Q = 650 m³/s, what is the correct calculated value of the average hydraulic radius for this reach? \newline A. 2.67 m; B. 5.34 m; C. 1.33 m; D. Cannot be calculated. \\
\addlinespace
Meteorology & Under neutrally stratified conditions, given a wind speed of 2.8 m/s at 2 m height in the surface layer and 4.9 m/s at 10 m height, the friction velocity $u^*$ is approximately ( ). \newline A. 0.2 m/s; B. 0.4 m/s; C. 0.5 m/s; D. 0.8 m/s. \\
\addlinespace
Power System & In power system short-circuit calculations, if the system base power is set at 125 MVA and the short-circuit power provided by the generator unit is 228.35 MVA, what is the short-circuit power supplied by the system side? \newline A. 771.65 MVA; B. 875.35 MVA; \newline C. 950.65 MVA; D. 1028.35 MVA \\
\bottomrule
\footnotetext{Note: These example questions are translated from Chinese.}
\end{longtable}

\section{Prompts applied in this study.}\label{secA2}
\begin{longtable}{p{\textwidth}}
\caption{Question Generation Prompts}\label{tab:question-generate-prompt} \\
\toprule
\textbf{Content} \\
\midrule
You are a highly capable knowledge-question generation assistant. Please create
\textless number\textgreater\ \textless multiple-choice\textgreater\ questions strictly
based only on the knowledge points mentioned in the given passage. \\
\addlinespace
\textbf{Requirements:} \\
1. The options of each question should cover difficult, easily confused, or
high-discrimination points in the passage. \\
2. Avoid absolute terms such as “always” or “all”. \\
3. Neither the questions nor the explanations may include expressions like
“according to the formula/example in the passage”, “as shown in Figure xx”, or
“based on the passage’s equation/example”. The questions must stand alone and
be understandable without the passage. \\
4. The \textless number\textgreater\ questions should be distributed across
different sections of the passage, covering its core knowledge points. \\
5. Questions must focus on assessing \textless basic conceptual knowledge\textgreater,
similar in difficulty to the given examples. \\
\addlinespace
\textbf{Examples:} \\
\textless Question: For the same river reach, how does the propagation speed of
a large flood compare to that of a small flood? A. Faster; B. Slower; C. About
the same; D. Cannot be determined.\\
Answer: A \\
Analysis: …… \textgreater \\
\addlinespace
\textbf{Output format:} \\
Output the results in strict json format. \\
\bottomrule
\multicolumn{1}{p{\textwidth}}{\footnotesize
The content within \textless\ \textgreater\ brackets represents a replaceable input variable.} \\
\multicolumn{1}{p{\textwidth}}{\footnotesize
Note: This prompt has been translated from Chinese.} 
\end{longtable}

\begin{table}[htbp]
\caption{Model Evaluation Prompts}\label{tab: LLM eval prompt} 
\begin{tabular*}{\textwidth}{@{\extracolsep\fill}p{\textwidth}}
\toprule
\textbf{Content} \\
\midrule
\textbf{Response generation prompt:} \\
You are an expert in the Hydro-Science and Engineering domain. Please answer the following \textless single-choice\textgreater\ question from a professional perspective and provide all correct answer options. \\
Question: \textless Question\textgreater \\
\addlinespace
\textbf{Answer extraction prompt:} \\
You are a professional answer-analysis assistant. Extract the correct choice letters from the following \textless single-choice\textgreater\ question response. \\
1. Return only the choice letters, such as A, B, C, D.\\
2. For multiple-choice questions, arrange letters in alphabetical order, e.g., AB, ACD.\\
3. If no explicit option letters are present in the answer, return “Unrecognizable”.\\
4. Return letters only, without any other text or symbols.\\
\addlinespace
Response content: \textless Response Content\textgreater \\
\addlinespace
Please directly return the option letters: \\
\bottomrule
\end{tabular*}
\footnotetext{The content within \textless\   \textgreater\ brackets represents a replaceable input variable.}
\footnotetext{Note: This prompt has been translated from Chinese.}
\end{table}

\begin{table}[htbp]
\caption{Confidence Estimation Prompts}\label{tab: confidence estimation prompt} 
\begin{tabular*}{\textwidth}{@{\extracolsep\fill}p{\textwidth}}
\toprule
\textbf{Content} \\
\midrule
You are a professional answer evaluation expert. Please assess the confidence level of the given answer.\\
Question: \textless Question\textgreater\\
Question type: \textless Single-choice\textgreater.\\
Question: \textless Answer\textgreater \\
\addlinespace
Please provide a confidence score for this answer as an integer from 1 to 5, where 5 indicates the highest confidence and 1 the lowest.\\ 
Return only the integer score without any additional text.\\
\bottomrule
\end{tabular*}
\footnotetext{The content within \textless\   \textgreater\ brackets represents a replaceable input variable.}
\footnotetext{Note: This prompt has been translated from Chinese.}
\end{table}

\clearpage
\section{Model performance statistics}\label{secA3}
\begin{table}[htbp]
\caption{Performance of LLMs on the Hydro-SE Bench}\label{tab:llm_performance}
\begin{tabular*}{\textwidth}{@{\extracolsep\fill}lcccc}
\toprule
\textbf{Model Name} & \textbf{\makecell{Accuracy on \\ question A}} & \textbf{\makecell{Accuracy on \\ question B}} & \textbf{\makecell{Accuracy on \\ question C}} & \textbf{Overall Accuracy} \\
\midrule
DeepSeek-V3.2-Exp & 0.809 & 0.770 & 0.826 & 0.796 \\
DeepSeek-R1 & 0.818 & 0.765 & 0.806 & 0.794 \\
Kimi-K2-0905 & 0.804 & 0.744 & 0.769 & 0.773 \\
GLM-4.5 & 0.792 & 0.741 & 0.797 & 0.772 \\
GPT-5 & 0.774 & 0.734 & 0.803 & 0.763 \\
Grok-4 & 0.775 & 0.716 & 0.807 & 0.757 \\
Qwen-Plus & 0.781 & 0.699 & 0.770 & 0.746 \\
Gemini-2.5-Pro & 0.767 & 0.691 & 0.781 & 0.738 \\
Llama-4 (Maverick) & 0.773 & 0.701 & 0.741 & 0.738 \\
Claude-4.5 (Sonnet) & 0.770 & 0.699 & 0.744 & 0.736 \\
GLM-4-32B-0414 & 0.727 & 0.630 & 0.677 & 0.679 \\
Qwen2.5-32B-Instruct & 0.721 & 0.629 & 0.549 & 0.653 \\
Qwen2.5-72B-Instruct & 0.698 & 0.628 & 0.593 & 0.651 \\
Qwen2.5-7B-Instruct & 0.635 & 0.516 & 0.436 & 0.551 \\
Llama-3-70B-Instruct & 0.584 & 0.493 & 0.421 & 0.518 \\
Llama-3-8B-Instruct & 0.462 & 0.392 & 0.319 & 0.408 \\
\bottomrule
\end{tabular*}
\footnotetext{Question A: basic conceptual knowledge; question B: engineering applications; question C: reasoning and calculation.}
\end{table}

\begin{figure}
    \centering
    \includegraphics[trim=1cm 0cm 1cm 5cm, clip, width=1.0\linewidth]{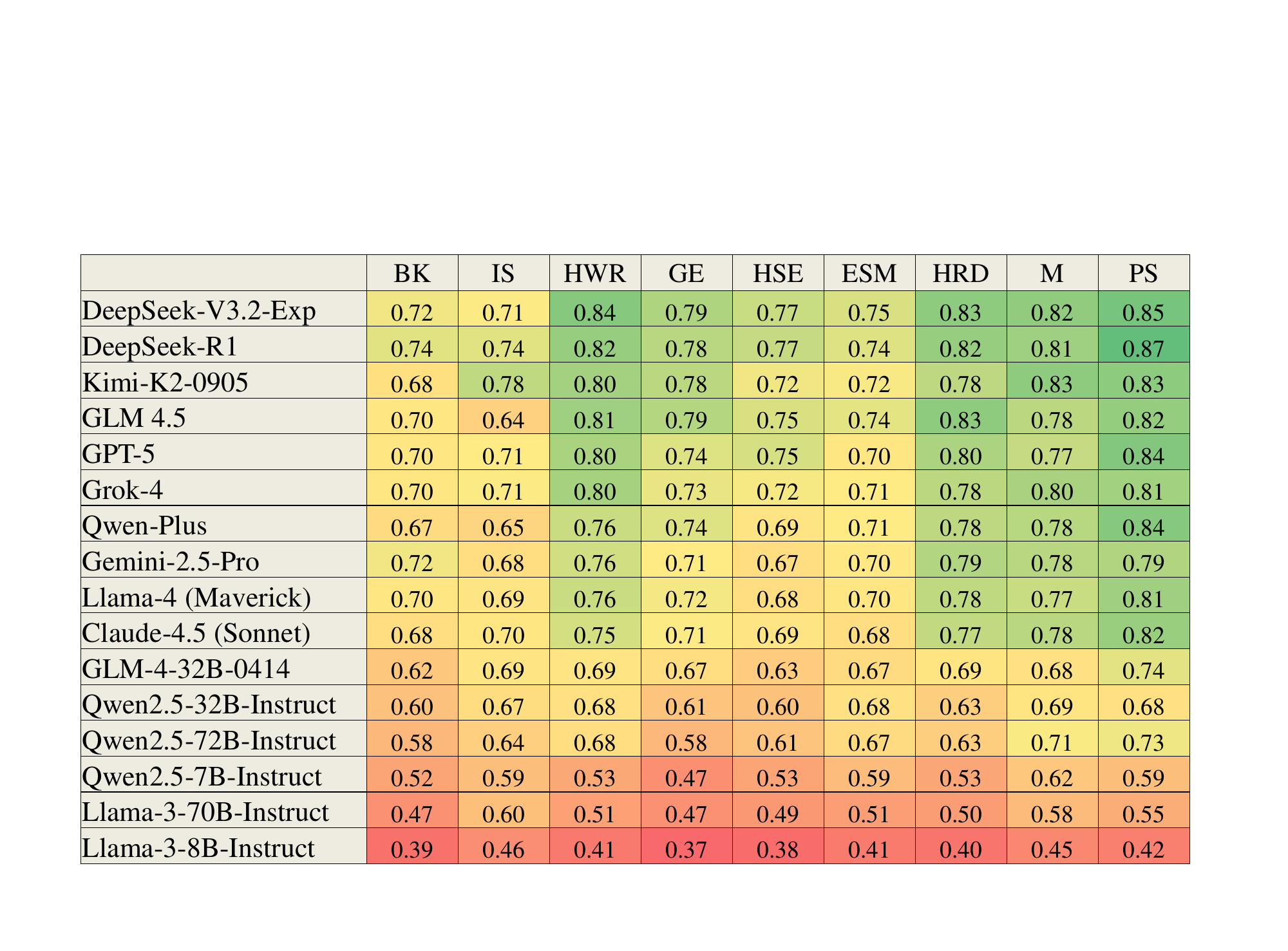}
    \caption{\textbf{Performance of LLMs on Hydro-SE Bench across nine subfields.} (BK: Background Knowledge; IS: Industry Standard; HWR: Hydrology and Water Resources; GE: Geotechnical Engineering; HSE: Hydraulic Structures and Equipment; ESM: Engineering Safety and Management; HRD: Hydraulics and River Dynamic; M: Meteorology PS: Power System)}
    \label{fig:Num_sub_perform}
\end{figure}

\begin{table}[ht]
\caption{LLM Token Consumption on the Hydro-SE Bench}\label{tab:llm_token}
\begin{tabular*}{\textwidth}{@{\extracolsep\fill}lclc}
\toprule
\textbf{Model Name} & \textbf{Token Consumption} & \textbf{Model Name} & \textbf{Token Consumption}  \\
\midrule
DeepSeek-V3.2-Exp & 4.057 M & Llama-4 (Maverick) & 4.798 M \\
DeepSeek-R1 & 8.296 M & Claude-4.5 (Sonnet) & 4.465 M \\
Kimi-K2-0905 & 2.707 M & GLM-4-32B & 3.733 M \\
GLM-4.5 & 14.70 M & Qwen-2.5-72B & 3.005 M \\
GPT-5 & 6.334 M & Qwen-2.5-32B & 2.329 M \\
Grok-4 & 16.284 M & Qwen-2.5-7B & 2.676 M \\
Qwen-Plus & 5.959 M & Llama-3-70B & 2.228 M \\
Gemini-2.5-Pro & 16.278 M & Llama-3-8B & 2.486 M \\

\bottomrule
\end{tabular*}
\footnotetext{M: Million}
\end{table}
\end{appendices}
\end{document}